\documentclass[sigconf]{acmart}
\AtBeginDocument{%
  }

\usepackage[]{hyperref}
\usepackage{multirow}
\usepackage{makecell}
\usepackage{booktabs}
\usepackage{filecontents}
\usepackage{colortbl}
\usepackage{xcolor}
\usepackage{tikz}
\usetikzlibrary{calc}
\usepackage{wrapfig}
\usepackage{array}
\usepackage[caption=false,font=normalsize,labelfont=sf,textfont=sf]{subfig}
\usepackage{textcomp}
\usepackage{stfloats}
\usepackage{url}
\usepackage{verbatim}
\usepackage{graphicx}
\usepackage{amsmath,amsfonts}
\usepackage{algorithmic}
\usepackage{algorithm}

\usepackage{array}

\DeclareMathOperator*{\argmin}{arg\,min}

\setcopyright{acmlicensed}
\copyrightyear{2018}
\acmYear{2018}
\acmDOI{XXXXXXX.XXXXXXX}
\acmConference[Conference acronym 'XX]{Make sure to enter the correct
  conference title from your rights confirmation email}{June 03--05,
  2018}{Woodstock, NY}
\acmISBN{978-1-4503-XXXX-X/2018/06}

\begin{document}

\title{DistrAttention: An Efficient and Flexible Self-Attention Mechanism on Modern GPUs}

 %\author{Haolin Jin}
% \affiliation{%

\author{Haolin Jin}
 \affiliation{%
   \institution{Shandong University}
   \country{China}
 }

 \author{Mengbai Xiao}\thanks{*Corresponding author: xiaomb@sdu.edu.cn}
 \authornotemark[1]
 \affiliation{%
   \institution{Shandong University}
   \country{China}
 }

 \author{Yuan Yuan}
 \affiliation{%
   \institution{Shandong University}
   \country{China}
 }

 \author{Xiao Zhang}
 \affiliation{%
   \institution{Shandong University}
   \country{China}
 }

 \author{Dongxiao Yu}
 \affiliation{%
   \institution{Shandong University}
   \country{China}
 }

 \author{Guanghui Zhang}
 \affiliation{%
   \institution{Shandong University}
   \country{China}
 }

 \author{Haoliang Wang}
 \affiliation{%
   \institution{Adobe}
   \country{USA}
 }

%  \author{Lars Th{\o}rv{\"a}ld}
%  \affiliation{%
%    \institution{The Th{\o}rv{\"a}ld Group}
%    \city{Hekla}
%    \country{Iceland}}
%  \email{larst@affiliation.org}

%  \author{Valerie B\'eranger}
%  \affiliation{%
%    \institution{Inria Paris-Rocquencourt}
%    \city{Rocquencourt}
%    \country{France}
%  }

%  \author{Aparna Patel}\thanks{∗Corresponding author: xiaomb@sdu.edu.cn
% }
%  \authornotemark[1]
%  \affiliation{%
%   \institution{Rajiv Gandhi University}
%   \city{Doimukh}
%   \state{Arunachal Pradesh}
%   \country{India}}

%  \author{Huifen Chan}
%  \affiliation{
%    \institution{Tsinghua University}
%    \city{Haidian Qu}
%    \state{Beijing Shi}
%    \country{China}}

% \affiliation{%

% \affiliation{%

% \affiliation{%

\renewcommand{\shortauthors}{Trovato et al.}

\begin{abstract}
  The Transformer architecture has revolutionized deep learning, delivering the state-of-the-art performance in areas such as natural language processing, computer vision, and time series prediction. However, its core component, self-attention, has the quadratic time complexity relative to input sequence length, which hinders the scalability of Transformers. The exsiting approaches on optimizing self-attention either discard full-contextual information or lack of flexibility. In this work, we design DistrAttention, an effcient and flexible self-attention mechanism with the full context. DistrAttention achieves this by grouping data on the embedding dimensionality, usually referred to as $d$. We realize DistrAttention with a lightweight sampling and fusion method that exploits locality-sensitive hashing to group similar data. A block-wise grouping framework is further designed to limit the errors introduced by locality sensitive hashing. By optimizing the selection of block sizes, DistrAttention could be easily integrated with FlashAttention-2, gaining high-performance on modern GPUs. We evaluate DistrAttention with extensive experiments. The results show that our method is 37\% faster than FlashAttention-2 on calculating self-attention. In ViT inference, DistrAttention is the fastest and the most accurate among approximate self-attention mechanisms. In Llama3-1B, DistrAttention still achieves the lowest inference time with only 1\% accuray loss. 
\end{abstract}



\if 0

\fi


\received{20 February 2007}
\received[revised]{12 March 2009}
\received[accepted]{5 June 2009}

\maketitle

%% Body

\section{Introduction}
\label{sec:introduction}
%-------------------------------------------------------------------------------
The emergence of the Transformer~\cite{vaswani2017attention} has significantly transformed the field of deep learning. 
% A key innovation of transformers is the introduction of the self-attention mechanism, which enables parallel evaluation of each token in the input sequence, thereby eliminating the sequential dependencies inherent in recurrent neural networks such as LSTM~\cite{hochreiter1997long}. This parallelism allows Transformers to fully leverage the capabilities of modern SIMD hardware accelerators like GPUs and TPUs, facilitating the training of models on datasets of unprecedented scale. 
By using self-attention, Transformer has shown unprecedented performance improvements across various domains, including natural language processing~\cite{brown2020language,yang2019xlnet,raffel2020exploring,kenton2019bert}, computer vision~\cite{han2023flatten,alexey2020image,carion2020end}, and time series forecasting~\cite{zhou2021informer}.
% , compared to the traditional sequence-to-sequence model~\cite{sutskever2014sequence} originally proposed for machine translation. 
% With the continuous growth of data, the demand for more efficient and powerful models has increasingly intensified, making the transformer architecture a focal point of current research. 
Despite its success, applications based on the Transformer architecture still face several challenges, one of which is the quadratic computational complexity with respect to input sequence length, limiting the ability to handle extremely long sequences or real-time applications.

A wide variety of Transformer optimizations have been proposed to reduce the quadratic complexity of the attention computation. Among common optimization methods, sparse attention~\cite{han2025agent,beltagy2020longformer,liu2016star}, which focuses on a subset of positions in the input tokens, can lead to a loss of important contextual information and inconsistent representations of the same input across different contexts. This results in suboptimal output, particularly for tasks that require capturing long-range dependencies or fine-grained details, and potentially degrades model robustness and generalization. Linear attention mechanisms~\cite{guo2024slab,you2024linear,han2024bridging} offer significant advantages in terms of computational efficiency and scalability by processing each token independently or in a fixed-size window. Since the interaction is limited to adjacent tokens only, linear attention undermines the effectiveness of a model in representing complex interactions across the entire sequence, such as language modeling or document-level translation. Quantization techniques~\cite{tang2024survey,lang2024comprehensive,du2024model} reduce the precision of model parameters to accelerate inference, but this thread of methods lacks flexibility: one quantized model makes a fixed trade-off between accuracy and latency. 
%we have to choose one among quarter-, half-, and single-precision floating point. Additional supports of hardware and libraries are also required. 
%which not only introduces additional complexity in terms of hardware and library support but is also challenging to achieve a good trade-off between accuracy and latency. 
%For certain hardware platforms, if there is a lack of support for low-precision computations, the benefits of quantization may be significantly diminished. 
%However, there is a notable scarcity of methods that can be applied to both inference and training across different types of models. Additionally, methods that effectively balance acceleration time with minimal precision loss are equally rare. This dual challenge underscores the need for more versatile and efficient optimization techniques. 

% In our paper, we observe that the computational complexity of attention in terms of visual complexity is quadratic with respect to the number of input tokens $n$, since each token needs to perform matrix multiplication with all other tokens. Consequently, many current approaches attempt to reduce the number of matrix multiplications per token. However, this inevitably leads to a loss of some complete token-to-token interaction information, thereby potentially diminishing the quality of the computed representations. 

In this paper, we propose distrAttention, an efficient self-attention mechanism with full contextual information and high flexibility. Instead of reducing input token length or quantizing parameters, we suggest to reduce the embedding dimensionality, which is commonly referred to as $d$ in literature. Changing $d$ does not change the attention matrix size so that we still have the full contextual information. By increasing the removed proportion of $d$, we could make flexible trade-offs between computation speedup and accuracy degradation. In DistrAttention, we do not directly remove columns and rows of matrices involved in the computation. Inspired by the distributive property of matrix multiplication, we group columns of $\mathbf{Q}$ and rows of $\mathbf{K}^\top$, which are two multiplied matrices in self-attention. By using $\mathbf{Q}$ column estimates and summing $\mathbf{K}^\top$ rows in groups, we calculate an approximate attention matrix with reduced multiplications. By arranging similar $\mathbf{Q}$ columns into a group, the attention matrix more approximates the ground truth. Thus, a lightweight sampling and fusion method based on locality-sensitive hashing (LSH) is designed to calculate an attention matrix approximation efficiently and accurately. To limit the error introduced by LSH, we further design a block-wise grouping mechanism, which fits the high-performance block-wise self-attention~\cite{dao-2024-flash} well. We also discuss that how to determine block sizes for achieving the optimal performance. We measure the accuray of DistrAttention with randomly generated $\mathbf{Q}$, $\mathbf{K}$, and $\mathbf{V}$, and the error is from 0.13\% to 0.07\% if we increase the sampling rate from 2 to 16. By comparing our method to FlashAttention-2, the self-attention is accelerated by up to 37\% with varying token and embedding length. In ViT inference, DistrAttention is the fastest (up to 8.7\% faster than the sencond place) and the most accurate (up to 8.1\% higer accuracy than the second place) among approximate self-attention mechanisms. In Llama3-1B, DistrAttention still achieves the lowest inference time (up to 15\% faster than the sencond place) and highest accuracy (up to 0.23\% higer accuracy than the second place). Our contributions are as follows:
\begin{itemize}
\item We propose an efficient and flexible self-attention mechanism, namely DistrAttention, that reduces computation at the embedding dimensionality $d$. 
\item We realize DistrAttention with a lightweight sampling and fusion method and a block-wise framework that fits the state-of-the-art implementation well.
\item We evaluate DistrAttention with extensive experiments, and the results show that our method is the fastest and most accurate among existing approximate attention mechanisms.
\end{itemize}

In Section 2, We discuss background and motivation, and in Section 3, We have provided a detailed explanation of how DistrAttention reduces the embedding dimensionality $d$ and the selection of block size. In Section 4, we design our experiments and conduct evaluations, while Section 5 discusses related work. Finally, Section 6 concludes our work.

\section{Background and Motivation}
\label{sec:motivation}

\subsection{The Standard Self-attention}
\label{sec:standard_attention}
% is the self-attention described in FlashAttention the standard version?
Self-attention serves as the core of Transformer, and it is designed to calculate pairwise ``attention'' between any two of the input tokens. Given the input matrices $\mathbf{Q}$, $\mathbf{K}$, and $\mathbf{V} \in \mathbb{R}^{N\times d}$, where $N$ is the sequence length and $d$ is the head dimension. The output is also a matrix $\mathbf{O} \in \mathbb{R}^{N\times d}$. Specifically, we have the computation as follows:
\[
  \mathbf{S} = \mathbf{Q}\mathbf{K}^\top \rightarrow \mathbf{P} = \text{softmax}(\mathbf{S}) \rightarrow \mathbf{O} = \mathbf{P}\mathbf{V},
\]
where $\text{softmax}(\cdot)$ is applied row-wise on $\mathbf{S}$. $\mathbf{S}$ ($\mathbf{P}$) is the (normalized) attention matrix composed of the pairwise $N^2$ attention scores.

\begin{figure}
  \centering
  \includegraphics[width=0.9\linewidth]{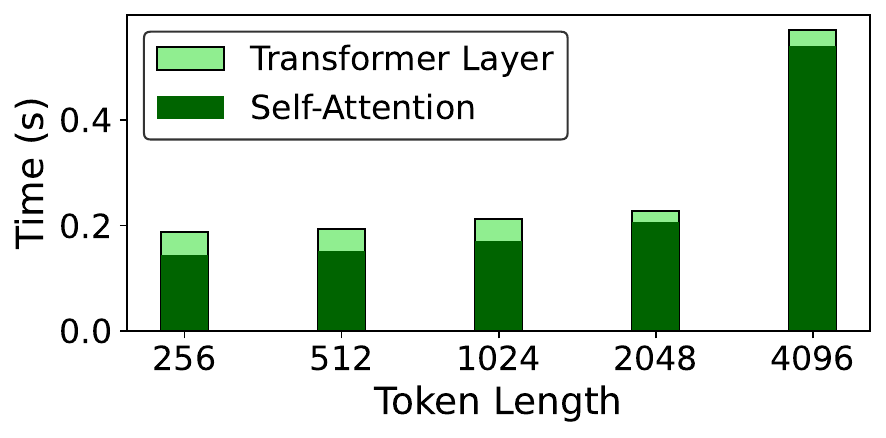}
  \caption{Computation time of self-attention in Transformer. The results are profiled by running a layer of Llama2-7B model, where $d$ is 64.}
  \label{fig:layer_time}
\end{figure}

Calculating self-attention has the complexity of $O(N^2)$, and it dominates the computation of a transformer block. We profile the computation time of a transformer layer in a Llama2-7B model~\cite{llama2_7b} on our testbed (See Section~\ref{sec:evaluation} for details), and report the results in Figure~\ref{fig:layer_time}. We can see that as the token length increases, the self-attention takes a larger proportion in terms of computation time. When 4K tokens are fed, 94\% time serves to compute self-attention.

\begin{figure}
  \centering
  \includegraphics[width=0.9\linewidth]{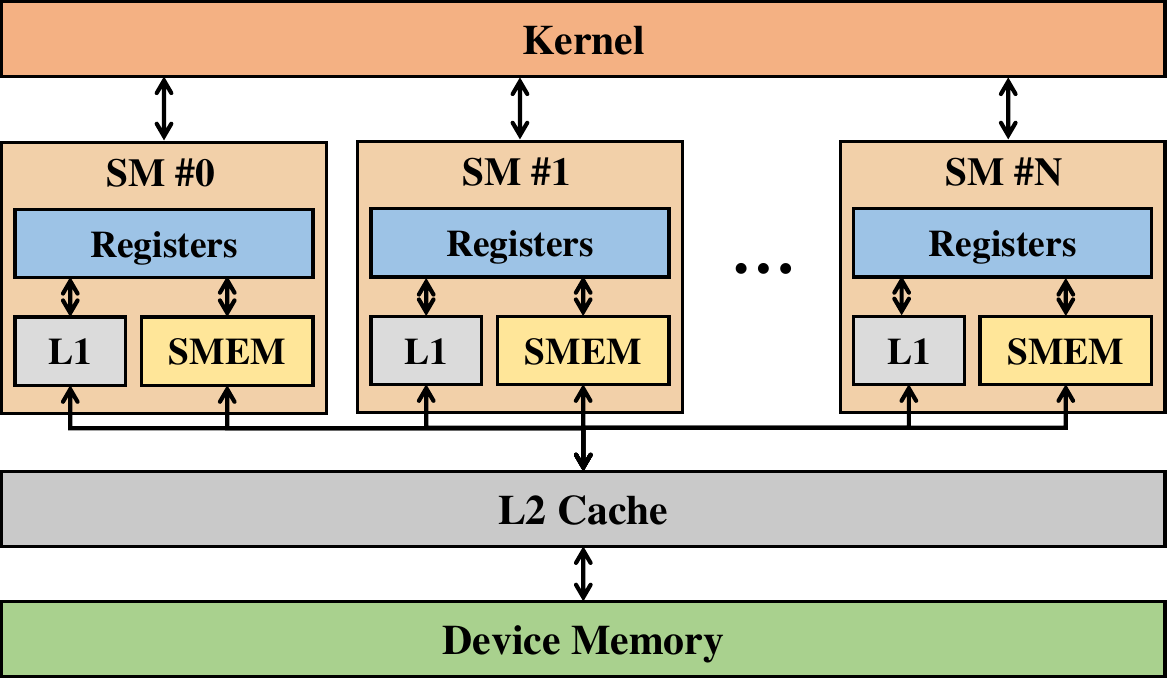}
  \caption{The memory system of a typical Nvidia GPU. SMEM stands for shared memory.}
  \label{fig:gpu_arch}
\end{figure}

\subsection{Self-attention on Modern GPUs}

Benefiting from the massively parallelism of GPUs, both training and inference of large-scale learning models become sustainable. Self-attention are mostly matrix operations, which could be substantially accelerated by modern GPUs equipped with Tensor cores.

\subsubsection{The GPU architecture}
On a modern GPU like Nvidia Ada architecture~\cite{nvidia2022adalovelace}, a number of Streaming Multiprocessors (SMs) are installed. An SM consists of CUDA cores, Tensor cores, and Ray-tracing cores. CUDA cores are used to execute instructions towards general computation tasks. Tensor cores and Ray-tracing cores are designed to specifically process tensor operations and ray-tracing tests, respectively. A GPU program is named a {\it kernel}. A kernel is organized to {\it threadblocks} that each is scheduled to an SM. Inside a threadblock, every 32 threads are grouped to {\it a warp}, and they are executed simultaneously, i.e., in the single-instruction-multiple-threads (SIMT) manner.

The memory architecture is also layered as at the CPU side. At the bottom layer, a unifying device memory is addressed by all SMs. There are two layers of cache, L2 cache is shared by SMs while each SM has its own L1 cache. The L1 cache could serve as a shared memory for threads in a threadblock. The threads are arranged individual register files during execution. Figure~\ref{fig:gpu_arch} shows the memory system on an Nvidia GPU.

\begin{figure*}[t]
  \centering
  \includegraphics[width=\textwidth]{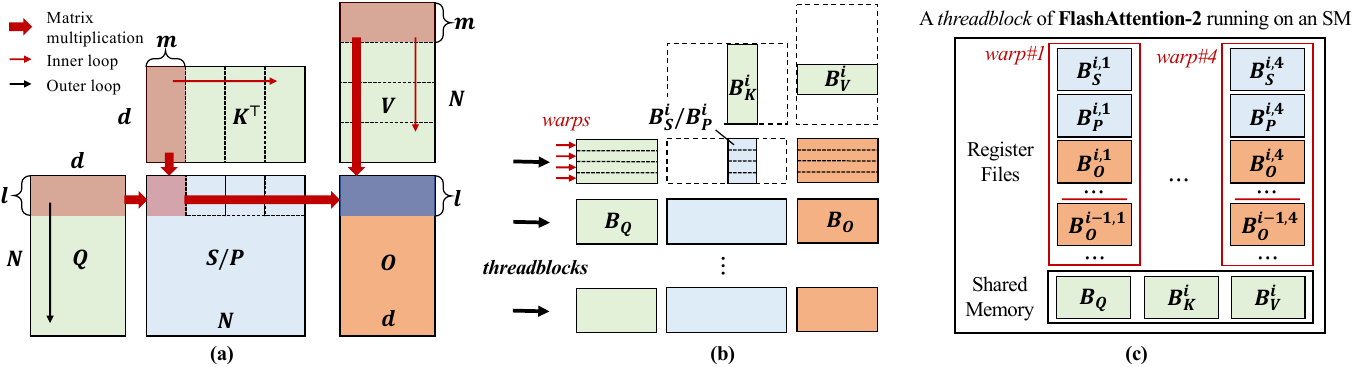}
  \caption{The illustration of FlashAttention-2~\cite{dao-2024-flash} on a modern GPU device.
    (a) The output matrix $\mathbf{Q}$ is calculated in a {\it double-loop}. The {\it outer loop} iterates over $\mathbf{Q}$ blocks to calculate $\mathbf{O}$ blocks. To calculate an $\mathbf{O}$ block, the {\it inner loop} iterates over $\mathbf{K}^\top$ and $\mathbf{V}$ blocks.
    (b) Calculating rows of $\mathbf{O}$ is embarrassingly parallel so that iterations of the outer loop are arranged to threadblocks. Inside a threadblock, a $\mathbf{Q}$ block is further splitted to sub-blocks and executed by warps.
    (c) The $\mathbf{Q}$, $\mathbf{K}$, and $\mathbf{V}$ blocks are loaded to shared memory, and the intermediate matrices are hold in register files. A few data structures in register files, like that for computing block-wise {\it softmax}, are ignored for clarity.
    }
  \label{fig:blockwise_flashattention}
\end{figure*}

\subsubsection{FlashAttention-2}
\label{sec:fa2}

Thanks to the high computation throughput of Tensors, the bottleneck of self-attention on GPU is memory I/Os~\cite{wolters-2024-memor-is}. FlashAttention~\cite{dao-2022-flash} and FlashAttention-2~\cite{dao-2024-flash} propose to reduce memory I/Os via block-wise self-attention and kernel fusion.

We use Figure~\ref{fig:blockwise_flashattention} to briefly introduce FlashAttention-2. $\mathbf{Q}$, $\mathbf{K}$, and $\mathbf{V}$ is splitted into blocks $\mathbf{B}_Q$, $\mathbf{B}_K$, and $\mathbf{B}_V$, respectively. The size of $\mathbf{B}_Q$ is $l\times d$ while the size of $\mathbf{B}_K$ and $\mathbf{V}_V$ is $m\times d$. $\mathbf{O}$ is calculated block-wise. By exploiting the shared memory and register files on the chip, FlashAttention-2 do not need 1) repeatedly read rows and columns of input matrices, and 2) materialize the intermediate matrix (like $\mathbf{S}$ and $\mathbf{P}$) to the global memory, effectively eliminating memory I/Os required by self-attention.

\subsection{Motivation}
To save execution time of self-attention, one approach is to prune tokens that have low attention scores with the others~\cite{wang2020linformer}, i.e., reducing $N$. But this method is not practical because it is hard to predict which tokens would generate low scores in advance. Moreover, directly removing tokens makes the context incomplete, severely impairing the output quality.

As reducing the dimensionality of $N$ degrades the accuracy of learning models with Transformers, we are interested in whether reducing the dimensionality of $d$ would result in acceleration. Reducing $d$ does not change the number of tokens but is a means of linearly removing information carried by embeddings. We expect this provides us with a more flexible method to tradeoff calculation speed with prediction accuracy of a learning model. Based on the introduction to block-wise self-attention in Section~\ref{sec:fa2}, changing $d$ to a small value not only incurs less computations (floating-point additions and multiplications) but also increases the size of matrix blocks loaded to shared memory, leading to less memory I/Os. Table~\ref{tab:reduce_d_fa} reports the time in milliseconds of running FlashAttention-2 with decreasing $d$, and we can observe that with the tokens increasing from 1K to 8K, halving $d$ results in 1.13x to 1.23x speedup. Hence, in this paper, we are looking for a faster self-attention by reducing $d$ without impairing the computation results.

\begin{table}[t]
  \centering
  \caption{The time ($\mu s$) of executing FlashAttention-2 with varying $N$ and $d$.}
  \label{tab:reduce_d_fa}
  \begin{tabular}{ccccc}
    \toprule
    $d$ & $N$=1024 & 2048 & 4096 & 8192 \\ 
    \midrule
    128 & 0.86 & 3.19 & 12.27 & 49.46 \\ 
    64 & 0.76 & 2.66 & 10.25 & 40.06 \\
    \bottomrule
  \end{tabular}
\end{table}

\section{Overview of DistrAttention}
\label{sec:overview}

%是否需要一个标准attention的global memory和share memory内存交换和时间关系的折线图，然后引出来针对的是I/O优化？

In this section, we introduce DistrAttention, a method of calculating self-attention with reduced $d$. Directly removing columns in $\mathbf{Q}$ and rows in $\mathbf{K}^\top$ results in an inaccurate output because of information loss. Inspired by the distributive property of matrix multiplication, we reduce $d$ by grouping similar $\mathbf{Q}$ columns. A lightweight sampling and fusion method based on locality-sensitive hashing (LSH) is developed to realize DistrAttention. To limit errors introduced by LSH, we further design a block-wise grouping method, which could be seamlessly integrated with the most efficient block-wise self-attention mechanism on GPUs, i.e., FlashAttention-2. In the end, we discuss how should we select the block sizes for maximizing performance.

% As the I/O between SRAM (Static Random-Access Memory) and HBM (High Bandwidth Memory) significantly influences the latency of the attention mechanism, reducing such I/O transactions is paramount to our research endeavors. First, in pursuit of comprehensive information capture and computational efficiency, we introduce a DistrAttention mechanism. This approach leverages the similarity along the column dimensions of Q and K to fusion the column-wise interactions of the Q and K matrices. By doing so, it enables our method to cohesively train and infer while being compatible with all existing attention optimizations. Second, based on the Flash Attention framework, we further optimize the I/O between SRAM and HBM, determining the optimal partitioning for $Q$ and $K$ computations. This refinement aims to precisely enhance the efficiency of attention computations. Our Synergistic Attention not only integrates seamlessly with current advancements but also pushes the boundaries by optimizing memory interactions, thus achieving a more accurate and efficient computation process.
%
%\begin{figure}
%    \centering
%    \includegraphics[width=0.8\linewidth]{figures/Memory_speed_1.pdf}
%    \caption{NAN NAN NAN NAN NAN NAN NAN}
%    \label{fig:I/O_time}
%\end{figure}
%

%-------------------------------------------------------------------------------
\subsection{Attention Matrix Approximation}
% -------------------------------------------------------------------------------
The attention matrix $\mathbf{S}$ serves as the core of self-attention, which is calculated by
\[
  \mathbf{S} = \mathbf{Q}\mathbf{K}^\top = \sum_{i=1}^d \mathbf{q}_i\mathbf{k}_i^\top,
\]
where $\mathbf{q}_i$/$\mathbf{k}_i^\top$ is a column/row of the matrix $\mathbf{Q}$/$\mathbf{K}^\top$. Inspired by the distributive property of matrix multiplication, we derive that
\begin{equation}
  \label{eq:qk_distr}
  \mathbf{S} = \mathbf{q}\sum_{i=1}^d \mathbf{k}_i^\top
  %\text{, or}\,\,\, (\sum_{i=1}^d \mathbf{q}_i)\mathbf{k}^\top
\end{equation}
if all columns are the same as $\mathbf{q}$. In such a case, the number of multiplications required is reduced to 1 from $d$ to calculate an element in the result matrix. 

\begin{figure}
  \centering
  \includegraphics[width=1\linewidth]{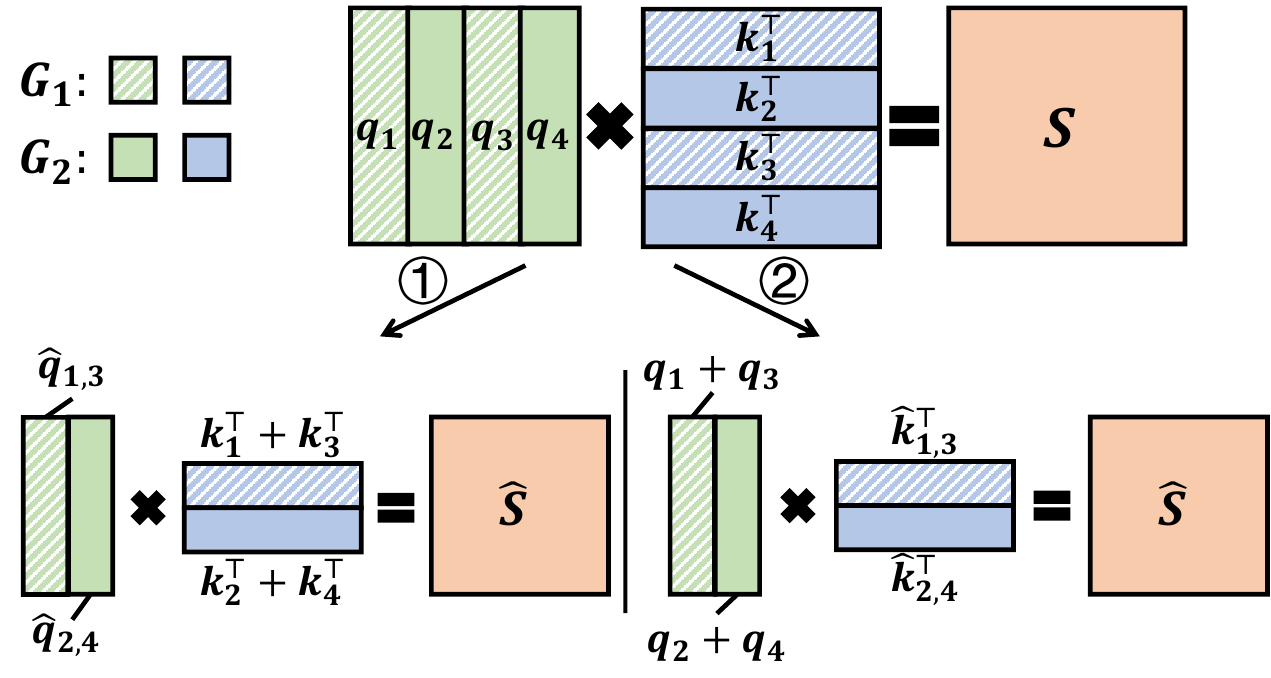}
  \caption{An example of calculating an attention matrix approximation $\hat{\mathbf{S}}$. The original $\mathbf{Q}$ and $\mathbf{K}^\top$ are splitted to two groups: $\{(\mathbf{q}_1, \mathbf{k}_1^\top), (\mathbf{q}_3, \mathbf{k}_3^\top)\}$ and $\{(\mathbf{q}_2, \mathbf{k}_2^\top), (\mathbf{q}_4, \mathbf{k}_4^\top)\}$. We either calculate $\hat{\mathbf{S}}$ by the estimates of $\hat{\mathbf{q}}_{1,3}$ and $\hat{\mathbf{q}}_{2,4}$ (the left side) or by the estimates of $\hat{\mathbf{k}}^\top_{1,3}$ and $\hat{\mathbf{k}}^\top_{2,4}$ (the right side). }
  \label{fig:distributive}
\end{figure}

In practice, the columns are hardly the same in $\mathbf{Q}$. But, we could approximate the attention matrix following the similar idea. Specifically, we group columns of $\mathbf{Q}$ to non-overlapping subsets $G_1, G_2, \cdots, G_k$, where $\bigcup_{j=1}^kG_j = \{1, 2, \cdots, d\}$. By using an estimate $\hat{\mathbf{q}}_j$ to replace all columns in $G_j$, we have
\begin{equation}
  \label{eq:approx_s}
  \hat{\mathbf{S}} = \sum_{j=1}^k\left( \hat{\mathbf{q}_j}\sum_{i\in G_j}\mathbf{k}_i^\top   \right), 
\end{equation}
where $\hat{\mathbf{S}}$ is an approximation of $\mathbf{S}$. Figure~\ref{fig:distributive} shows an example of grouping $\mathbf{q}$s to calculate an approximate $\mathbf{S}$. It is worth noting that the same operations could be applied on $\mathbf{K}^\top$ to get the approximation matrix $\hat{\mathbf{S}}$ as well.

To make $\hat{\mathbf{S}}$ an accurate approximation, we try minimizing $\| \hat{\mathbf{S}} - \mathbf{S} \|_1$, where $\| \cdot \|_1$ is the sum of the absolute values of the components.  This leads to an optimization problem formulated as 
\begin{equation}
  \label{eq:goal}
    \argmin_{G_1, G_2, \cdots, G_k} \sum_{j=1}^k\sum_{i\in G_j}\left\| \left(\hat{\mathbf{q}_j} - \mathbf{q}_i \right)\mathbf{k}_i^\top \right\|_1.
\end{equation}
When solving Equation~\ref{eq:goal}, if we allow $G_j$ to contain one column and use it as the estimate, it leads to a solution of  $\hat{\mathbf{S}}$ equaling $\mathbf{S}$, but it saves no multiplication. Thus, we are more likely to impose a constraint of $|G_j| > 1$ to balance the computation speedup and result accuracy.

In a learning model, $\mathbf{Q}$ and $\mathbf{K}$ vary across batches, transformer blocks, and heads. It is not practical to solve the optimization problem for each head as the self-attention is commonly calculated at the scale of sub-milliseconds. Thus, we propose a lightweight grouping strategy that effectively reduce computation without impairing the result accuracy.

\subsection{Lightweight Sampling and Fusion}
\label{sec:sampling_and_fusion}

To avoid time-costing solution on the optimization problem in Equation~\ref{eq:goal}, we first limit the solution space by imposing a constant group size $G^*$, i.e., a group contains $2, 4, \cdots$ columns. Then, we strive to arrange the similar columns into a group. For a group $G_j$, we select one of their $\{\mathbf{q}_i, i\in G_j\}$ columns as the estimate  $\hat{\mathbf{q}}_j$ ({\it sampling}) and sum all $\{\mathbf{v}_i, i\in G_j\}$ ({\it fusion}). Multiplying the sampled columns by the fused rows results in the final attention matrix approximation $\hat{\mathbf{S}}$. If the $\mathbf{Q}$ columns in each group are similar enough, the error of $\hat{\mathbf{S}}$ approaches 0.

\noindent{\bf Similar columns idenfication}. We identify the similar columns using locality-sensitive hashing (LSH). Given a column $\mathbf{q} \in \mathbb{R}^{N\times 1}$, we maps it to an integer. Specifically, $\mathbf{q}$ is projected into a low-dimensional space of $N'$ followed by a binarization step, where positive values are set to 1s and the others are set to 0s. Then, the binary sequence is used as the index to select a value from a table of Gray code at the size of $2^{N'}$. As a result, $\mathbf{Q} \in \mathbb{R}^{N\times d}$ is mapped to a set of hash values $\mathbf{Q}_H \in \mathbb{N}^{1\times d}$. The projection matrix is randomly generated in prior and $N'$ is set to 16 to match the tensor size commonly accepted by Tensor cores on commodity GPUs.

\begin{figure}
  \centering
  \includegraphics[width=1\linewidth]{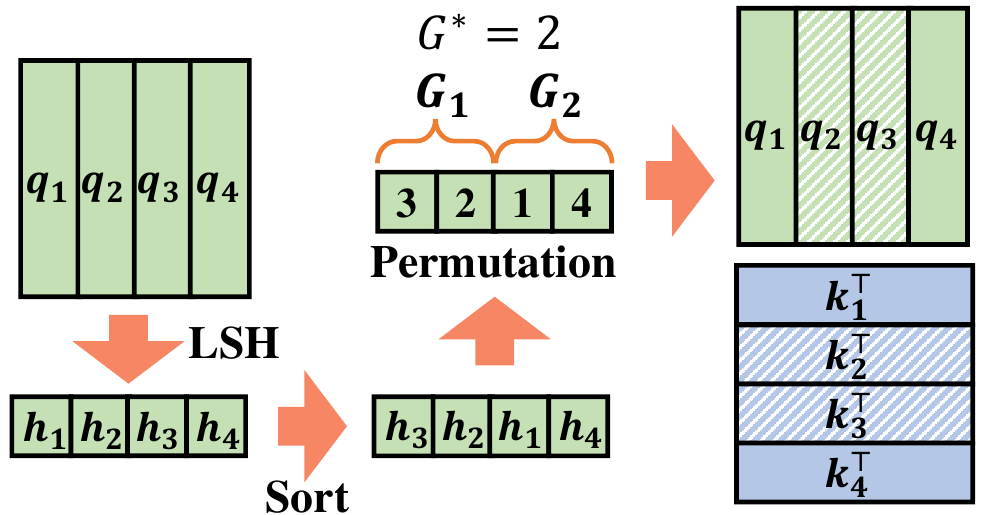}
  \caption{An example of grouping $\mathbf{q}$s with LSH. 4 columns of $\mathbf{Q}$ is hashed to 4 integer values $\{h_1, h_2, h_3, h_4\}$ that are further sorted to $\{h_3, h_2, h_1, h_4\}$. We then have an index permutation of $\{3, 2, 1, 4\}$. By imposing a group size $G^* = 2$, we know that $\mathbf{q}_3$ and $\mathbf{q}_2$ are in a group while $\mathbf{q}_1$ and $\mathbf{q}_4$ are in another group.}
  \label{fig:lsh_group}
\end{figure}

Due to the property of LSH, a small difference between two hash values indicates the $\mathbf{q}$s are also close in the $N$-dimensional space with a high probability. So, we group $\mathbf{q}$s according to the hash values: The hash values are monotonically sorted and thus an index permutation of represented columns is generated. Following the permutation, every $G^*$ specified columns (in $\mathbf{Q}$) and rows (in $\mathbf{K}^\top$) are placed in a group. Figure~\ref{fig:lsh_group} shows an example of grouping columns and rows. With a generated permutation and a given $G^*$, it is trivial to carry out the sampling and fusion operation in each group, as well as to calculate the attention matrix approximation.

\subsection{Block-wise Grouping}
As $N$ grows, using the hash values to indicate the closeness of high-dimensional vectors is error-prone. We propose to split $\mathbf{Q}$ and $\mathbf{K}^\top$ into blocks and always carry out LSH from a limited dimension. More importantly, this also fits the computation paradigm of block-wise self-attention introduced in Section~\ref{sec:fa2}.

\begin{figure}
  \centering
  \includegraphics[width=1\linewidth]{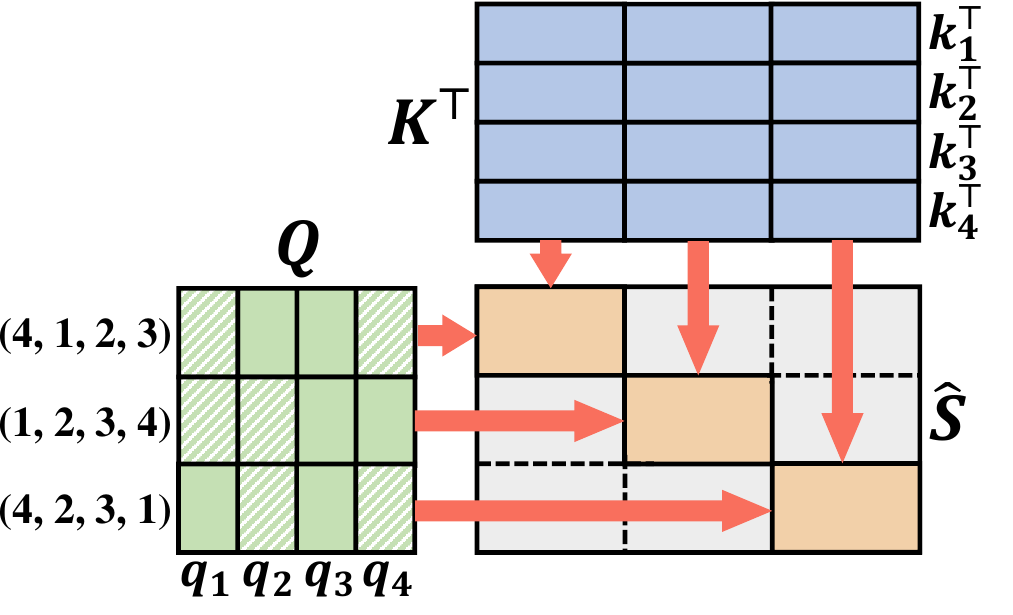}
  \caption{An example of block-wise grouping. $\mathbf{Q}$ and $\mathbf{K}^\top$ are splitted into 3 blocks, though both the size and number of $\mathbf{Q}$ and $\mathbf{K}^\top$ blocks do not have to be the same. The $\mathbf{Q}$ blocks are grouped individually, and the rows of a $\mathbf{K}$ block are grouped depending on which $\mathbf{Q}$ block is multiplied by. The complete $\hat{\mathbf{S}}$ is calculated using all 9 pairs of $\mathbf{Q}$ and $\mathbf{K}^\top$ blocks.}
  \label{fig:blockwise_grouping}
\end{figure}

As shown in Figure~\ref{fig:blockwise_grouping}, we carry out the block-wise self-attention. Given a block of $\mathbf{Q}$ and a block of $\mathbf{K}^\top$, a block of $\hat{\mathbf{S}}$ is calculated by the same method described in Section~\ref{sec:sampling_and_fusion}. As long as all $\hat{\mathbf{S}}$ blocks have been computed, we have a full attention matrix approximation. It is worth noting that across $\mathbf{Q}$ blocks, we generate different permutations, and the grouping of rows in $\mathbf{V}$ blocks vary depending on the $\mathbf{Q}$ block. Applying multiple permutations results in a lower error compared to using a unique grouping strategy over the complete $\mathbf{Q}$ and $\mathbf{K}^\top$. As the number of $\mathbf{Q}$ blocks increases, more permutations are generated and thus a more accurate $\hat{\mathbf{S}}$ is computed.

When calculating $\hat{\mathbf{S}}$ with a double-loop, we iterate over $\mathbf{Q}$ blocks in the outer loop and over $\mathbf{K}^\top$ blocks in the inner loop. In this way, the same group that is determined by the column similarity of a $\mathbf{Q}$ block could be used to consecutively calculate a row of $\hat{\mathbf{S}}$ blocks. This is why we choose to sample on the $\mathbf{Q}$ columns instead of $\mathbf{K}^\top$ rows. In Equation~\ref{eq:qk_distr}, we may also approximate the attention matrix $\mathbf{S}$ via $\left(\sum\mathbf{q}_i\right)\mathbf{k}^\top$, but this requires re-loading or re-calculating the permutation in every iteration step. Switching $\mathbf{Q}$ and $\mathbf{K}$ in the double-loop structure is also not preferred since this results in a larger intermediate $\mathbf{O}$ which has inferior performance.% \footnote{This is a design in FlashAttention-1 that has been discarded in FlashAttention-2.}  

\subsubsection{Block size selection}
Selecting a small block size reduces the error of $\hat{\mathbf{S}}$, but it might impair performance of block-wise self-attention. As discussed in Section~\ref{sec:fa2}, smaller blocks incurs more memory I/Os thus longer execution time. To select the proper block sizes $l$ in terms of $\mathbf{Q}$ and $m$ in terms of $\mathbf{K}^\top$ and $\mathbf{V}$, we first analyze the I/O complexity of block-wise self-attention. 

% In this section, we attempt to analyze how I/Os are affected by the parameters.
According to the computation process presented in Section~\ref{sec:fa2}, we have
\[
  I(l, m) = \frac{N}{l}(ld + 2Nd + ld),
\]  
where $I(\cdot)$ is the overall I/O times of $l$ and $m$. $N/l$ indicates the number of $\mathbf{O}$ blocks to be computed. To compute an $\mathbf{O}$ block, we read a $\mathbf{Q}$ block ($ld$), read the complete $\mathbf{K}^\top$ and $\mathbf{V}$ ($2Nd$), and write an $\mathbf{O}$ block ($ld$). From this, we could conclude that the memory I/Os are independent of $m$, and a larger $l$ always leads to less I/O times. So, to minimize memory I/Os, it is encouraged to set $m$ to 1 so that the shared memory could accommodate a larger $\mathbf{Q}$ block.

% The blockwise matrix multiplication is illustrated in Figure~\ref{fig:blockwise_flashattention}. To calculate a  $l \times m$ block of $\mathbf{S}$, one has to load a $l \times d$ block of $\mathbf{Q}$ and a $m \times d$ block of $\mathbf{K}^\top$. Then, the block of $\mathbf{P}$ at the same size as $\mathbf{S}$ is calculated, which is further multiplied by a $m\times d$ block of $\mathbf{V}$. It is worth noting that the result matrix is not a block of $\mathbf{O}$ yet. A $l \times d$ block of $\mathbf{O}$ is derived by accumulating the result matrix of multiplying $\mathbf{P}$ blocks and $\mathbf{V}$.

% As $l$ and $m$ is constrained by the size of shared memory $M_s$, we have
% \begin{equation}
%   \label{eq:shared_constraint}
%   M_s \ge w(ld+2md),
% \end{equation}
% where $w$ is the size of an element in the matrices. 

But after experimenting a large number of $l$ and $m$ configurations, we notice that other factors also affects the selection of $l$ and $m$. The fast computation of self-attention on GPUs is rooted in the co-shipped Tensor cores, which accept input matrices at a fixed-size. Setting $m$ to 1 significantly lowers the computation throughput of Tensor cores. Thus, $l$ and $m$ are expected to be
\begin{equation}
  \label{eq:minimum}
  l, m = nN',
\end{equation}
where $N'$ is the fixed-size accepted by Tensor cores and $n = 1, 2, \cdots$.

Second, the execution of a warp switches between CUDA cores and Tensor cores. Thus, the number of warps excuted on an SM is expected to be twice of the Tensor cores for saturating the computation capability. So, we have $W_s \ge 2N_T$, where $W_s$ and $N_T$ are the number of warps and Tensor cores on an SM, respectively. This is expanded to
\begin{equation}
  \label{eq:tensor_util}
  W_b \cdot \frac{M_s}{w(ld+2md)} \ge 2N_T,
\end{equation}
where $W_b$ is the number of warps in a threadblock, $M_s$ is the size of shared memory, and $w$ is the size of an element in the matrices. $\frac{M_s}{w(ld+2md)}$ represents the number of threadblocks running on an SM.

\begin{table}[t]
  \centering
  %\footnotesize
  \caption{The selection of ($l$, $m$) in FlashAttention-2 ({\it flash}), our method ({\it ours}), and the optimal configuration ({\it best}) on Nvidia RTX 4090, RTX 3090, and L40}
  \label{tab:lm_selection}
  \begin{tabular}{ccccc}
    \toprule
    \bf{GPU}  & \bf{Method} & $d$=32 & $d$=64 & $d$=128 \\
    \midrule
    \multirow{3}{*}{\bf 4090} & {\it flash} & {(128, 128)} & {(128, 128)} & {(128, 32)} \\ 
                    & {\it ours} & {(256, 64)} & {(128, 128)} & {(128, 32)} \\ 
                    & {\it best} & {(128, 128)} & {(128, 128)} & {(128, 32)} \\ 
    \midrule
    {\multirow{3}{*}{\bf 3090}} & {\it flash} & {(128, 128)} & {(128, 128)} & {(128, 32)} \\ 
                    & {\it ours} & {(256, 64)} & {(128, 128)} & {(128, 32)} \\ 
                    & {\it best} & {(128, 128)} & {(128, 128)} & {(128, 32)} \\ 
    \midrule
    {\multirow{3}{*}{\bf L40}} & {\it flash} & {(128, 128)} & {(128, 128)} & {(128, 32)} \\ 
                    & {\it ours} & {(256, 64)} & {(128, 128)} & {(128, 32)} \\ 
                    & {\it best} & {(256, 64)} & {(128, 128)} & {(128, 32)} \\ 
    \bottomrule
  \end{tabular}
\end{table}

Besides, a larger $m$ is always preferred because it has less iteration overhead (inner-warp) and scheduling overhead (inter-warps). As a result, we always maximize $l$ then $m$ while following the constraints in Equation~\ref{eq:minimum} and Equation~\ref{eq:tensor_util}. With these rules, we calculate the selection of $l$ and $m$ under different $d$, and compare it with the hardcoded configurations of FlashAttention-2 in Table~\ref{tab:lm_selection}. In the table, the optimal configuration is discovered by testing all legal $l$ and $m$ combinations. We could observe that when $d$ is 64 and 128, our method finds the same configuration as FlashAttention-2, and both are the optimal configuration. When $d$ is 32, we choose a different $l$ and $m$ combination, but the performance gap is less than 1\% no matter ours or FlashAttention-2 is tested to be the best.

\begin{figure*}[t]
  \centering
  \includegraphics[width=1.0\linewidth]{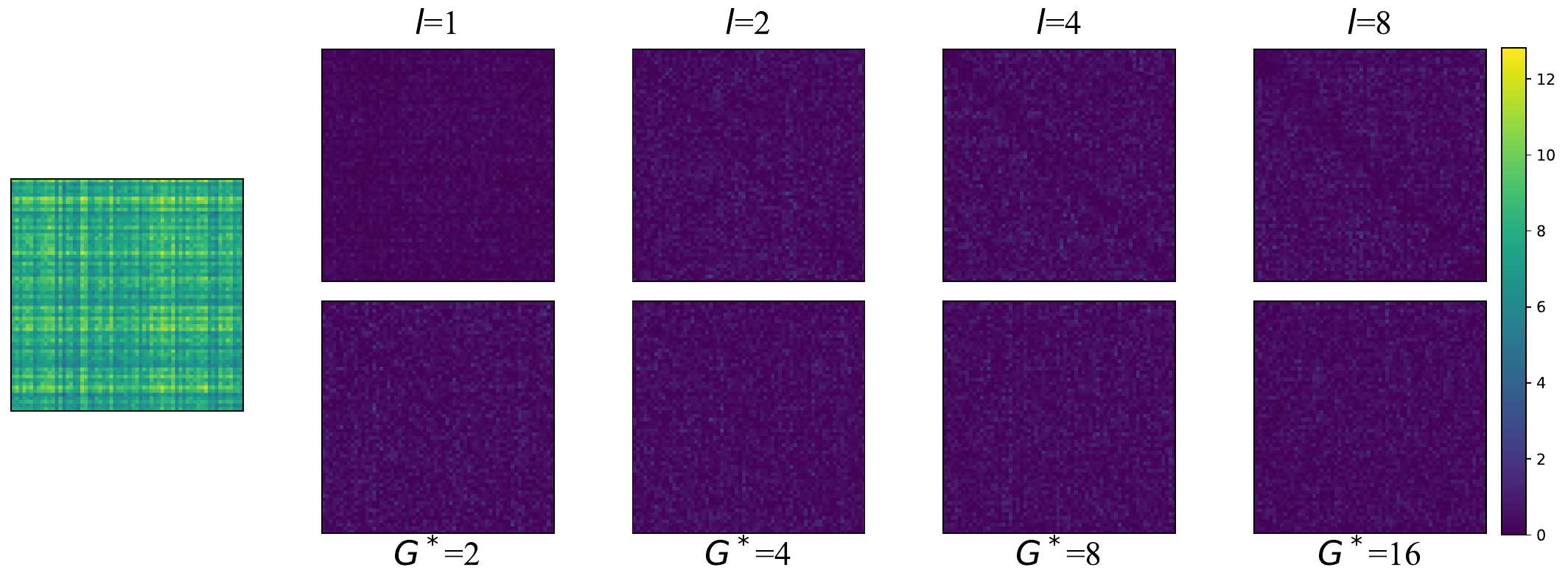}
  \caption{The errors of between $\hat{\mathbf{S}}$ and $\mathbf{S}$ on a pair of synthesized $\mathbf{Q}$ and $\mathbf{K}$, whose $N=64$ and $d=32$. We vary the block size $l$ and the sampling rate $G^*$ when $G^*$ is 2 and $l$ is 2, respectively. The original $\hat{\mathbf{S}}$ and $\mathbf{S}$ are shown at the left.
  %The errors between $\hat{\mathbf{S}}$ and $\mathbf{S}$ with varying sampling rate ($G^*$) and block size ($l$). Ground truth at the left side is the visualized $\mathbf{S}$. 
    % The errors between the approximated $\hat{\mathbf{S}}$ matrix calculated using different block grouping sizes and various column sampling rates, and the original $S$ matrix are showed. (a) shows the result of multiplying the unprocessed $Q$ and $K$ matrices. (b), (c), (d), and (e) represent the errors between the approximated $\hat{\mathbf{S}}$ matrix and the original S matrix, obtained by using different block sizes, applying grouped similarity, sampling columns of $Q$, and fusing columns of $K$ before multiplication. (f) illustrates the $\hat{\mathbf{S}}$ matrix obtained when using a block size of 2 and a sampling rate of 50\%. (g), (h), and (i) show the errors between the $\hat{\mathbf{S}}$ matrix and the original S matrix, processed with a block = 2 and varying sampling rates of 50\%, 25\%, 12.5\%, and 6.25\% for the Q and K matrices.
  }
  \label{fig:simility_chunk}
\end{figure*}

\section{Evaluation}
\label{sec:evaluation}

\subsection{Experimental Setup}
\paragraph{Testbed.} We evaluate our method and baselines on two servers. One is equipped with an NVIDIA RTX4090 and the other is installed an NVIDIA L40. Both machines are equipped with an Intel Xeon Platinum 8352V CPU, 1T main memory, and the operating system is Ubuntu 22.04.

% In our experiments, we validated the effectiveness of our optimization methods on both attention-based visual models and language models. We implemented the proposed model in python 3.8, our visual model experiments are carried out on the platform equipped with Nvidia RTX4090, an Intel Xeon Platinum 8352V CPU, 1T DDR memory, and the OS is Ubuntu 22.04, the language model experts are carried out on the platform equipped with Nvidia L40.

\paragraph{Baselines.} We compare our method with both exact attention methods and approximate ones. For exact attention, we use the standard attention mechanisms (Attn-Standard) those are by default deployed in target models, like ViT~\cite{pmlr-v139-touvron21a}, BERT~\cite{DBLP:journals/corr/abs-1810-04805}, and Llama 3~\cite{grattafiori2024llama}, and also use FlashAttention-2 (Flash2)~\cite{dao-2024-flash}. For approximate attention, the baselines include Primal attention (Primal)~\cite{chen2024primal}, Hyper attention (Hyper)~\cite{han2023hyperattention}, Flatten attention (Flatten)~\cite{han2023flatten}, and Hydra attention (Hydra)~\cite{bolya2022hydra}.

% \paragraph{Baseline.}We compare DistrAttention with three types of attention methods: (a) base attention, such as the inherit attention mechanism of Vit-base, BERT and LLaMA. (b) approximate attention methods, including Primal attention~\cite{chen2024primal}, Hyper attention~\cite{han2023hyperattention}, Flatten attention~\cite{han2023flatten}, Hydra attention~\cite{bolya2022hydra}. (c) Flash attention~\cite{dao-2024-flash}.

\paragraph{Workloads.} The different attention mechanisms are evaluated in both training and inference. We train the ViT model with five datasets, including ImageNet~\cite{deng2009imagenet}, CIFAR-100~\cite{Krizhevsky09learningmultiple}, CIFAR-10~\cite{Krizhevsky09learningmultiple}, iNaturalist 2018~\cite{van2018inaturalist}, and iNaturalist 2019~\cite{inaturalist-2019-fgvc6}, where the test sets are also contained. We also train the LLaMA3-1B model with the MathInstruct dataset and test it with MMLU-math. When evaluating inference workloads, we test the pre-trained ViT model on ImageNet and the pre-trained BERT model on the Policy dataset. %{\color{red}{Check the spelling of datasets}} 

%\paragraph{Training and Evaluation.}We conduct the extensive experiments for demonstrate the superior of DistrAttention from the perspective of training and inference directly. For training, we train and evaluate the vit model with five datasets, including ImageNet, Cifar100, Cifar10, Inat-comp2018 and Inat-comp2019. Furthermore, we train the LLaMA3-1B model with MathInstruct dataset and evaluate it with MMLU-math dataset. For inference directly, we evaluate ViT model on Imagenet and BERT on Policy dataset.

\subsection{The errors of $\hat{\mathbf{S}}$ on synthesized workloads}
Clearly, the effectiveness of DistrAttention is determined by how much the estimated $\hat{\mathbf{S}}$ is deviated from the original $\mathbf{S}$. We then first evaluate the errors introduced by our method on synthesized workloads. We randomly construct $\mathbf{Q}$ and $\mathbf{K}$ matrices ($N=64, d=64$), where an element is individually generated following a uniform possibility in $(0,1)$. Then, we calculate the errors between $\hat{\mathbf{S}}$ and $\mathbf{S}$ with varying configurations. The experiment is repeated 100 times. In the experiment, we vary the block size and the sampling rate. The block size ($l$) is selected among 1, 2, 4, and 8. The sampling rate ($G^*$) ranges in 2, 4, 8, and 16. When the block size (sampling rate) is varying, the sampling rate (block size) is fixed to 2. With different configurations, we calculate errors of $\hat{\mathbf{S}}$ from $\mathbf{S}$.

The statistical errors of attention matrix elements are reported in Table~\ref{tab:block_errs}, which includes the minimum ({\it min}), the maximum ({\it max}), and the average error ({\it mean}). The results show that with varying block sizes, the average error introduced ranges from 0.87\% to 0.9\%, and the maximum error changes from 3.4\% to 3.45\%. When we change the sampling rate, the average error is at least 0.87\% (at most 4.97\%), and the maximum error is at least 3.4\% (at most 16.5\%). Thus, the error of $\hat{\mathbf{S}}$ is more sensitive to the block size/sampling rate. We also visualize one of 100 experimental results in Figure~\ref{fig:simility_chunk}, where the errors of $\hat{\mathbf{S}}$ are hardly observed.

\if 0
\begin{table}[t]
\centering
\caption{Error Analysis of the Approximated Matrix $\hat{\mathbf{S}}$ and the Original Matrix $\mathbf{S}$ Under Different Block Sizes.}
\label{tab:block_errs}
%\renewcommand{\arraystretch}{1.5}
\begin{tabular}{cccccc}
\toprule
{} & block=1 & {2} & {4} & {8} & {$\mathbf{S}$} \\ 
\midrule
{min} & {0.0141} & {0.0001} & {0.0002} & {0.0009} & {7.7031} \\
%\makecell{max} & \makecell{0.0108} & \makecell{0.0470} & \makecell{0.0270} & \makecell{{0.0192} & \makecell{0.0224} \\ 
 {max} & {0.0108} & {0.047} & {0.027} & {0.0192} & {8} \\
 %{var} & {0.0005} & {0.0005} & {0.0004} & {0.0002} & {0.0224} \\
 {mean} & {0.0003} & {0.0133} & {0.009} & {0.0063} & {8.5} \\ 
 \bottomrule
\end{tabular}
\end{table}

\begin{table}[t]
\centering
\caption{Error Analysis of the Approximated Matrix $\hat{\mathbf{S}}$ and the Original Matrix $S$ Under Different Sampling Rate.}
\label{tab:sampling_rate}
%\renewcommand{\arraystretch}{1.5}
\begin{tabular}{cccccc}
\toprule
{} & Sampling rate=2 & {4} & {8} & {16} & {$\mathbf{S}$} \\
\midrule
{min} & {0.0008} & {0.0001} & {0.0002} & {0.0002} & {7.7031} \\
%\makecell{max} & \makecell{0.0108} & \makecell{0.0470} & \makecell{0.0270} & \makecell{{0.0192} & \makecell{0.0224} \\ 
 {max} & {0.0433} & {0.0422} & {0.0363} & {0.02} & {8} \\
 %{var} & {0.0006} & {0.0008} & {0.0007} & {0.0002} & {0.0224} \\
 {mean} & {0.0117} & {0.0142} & {0.0107} & {0.0061} & {8.5} \\ 
 \bottomrule
\end{tabular}
\end{table}
\fi

\begin{table}[t]
\centering
\caption{Error Analysis of the Approximated Matrix $\hat{\mathbf{S}}$ and the Original Matrix $\mathbf{S}$ Under Different Block Sizes. The data in the table represent the percentage of the current error relative to the true value.}
\label{tab:block_errs}
\begin{tabular}{ccccc}
\toprule
{} & $l$=1 & {$l$=2} & {$l$=4} & {$l$=8} \\ 
\midrule
{min} & {4E-4} & {2E-3} & {2E-3} & {1E-3}  \\
%\makecell{max} & \makecell{0.0108} & \makecell{0.0470} & \makecell{0.0270} & \makecell{{0.0192} & \makecell{0.0224} \\ 
 {max} & {3.45} & {3.4} & {3.4} & {3.4} \\
 %{var} & {0.0034} & {0.0036} & {0.0034} & {0.0029}  \\
 {mean} & {0.9} & {0.87} & {0.87} & {0.87}  \\ 
 \bottomrule
\end{tabular}
\end{table}

\begin{table}[t]
\centering
\caption{Error Analysis of the Approximated Matrix $\hat{\mathbf{S}}$ and the Original Matrix $S$ Under Different Sampling Rate. The data in the table represent the percentage of the current error relative to the true value.}
\label{tab:sampling_rate}
\begin{tabular}{cccccc}
\toprule
{} &$G^*$=2 & {$G^*$=4} & {$G^*$=8} & {$G^*$=16}  \\
\midrule
{min} & {2E-3} & {4E-2} & {7E-2} & {6E-3}  \\
%\makecell{max} & \makecell{0.0108} & \makecell{0.0470} & \makecell{0.0270} & \makecell{{0.0192} & \makecell{0.0224} \\ 
 {max} & {3.4} & {5.4} & {9.25} & {16.5}  \\
 %{var} & {0.0036} & {0.0096} & {0.001} & {0.009}  \\
 {mean} & {0.87} & {1.73} & {2.48} & {4.96}  \\ 
 \bottomrule
\end{tabular}
\end{table}

\begin{table*}[t]
\caption{Performance comparison of models trained on ImageNet and Fine-Tuned Using Different Methods Across Various Datasets. $r1$ and $r2$ denote the fine-tuning processes using different learning rates.}
\label{tab:vis_method}
\centering
\renewcommand{\arraystretch}{1.0}
\scalebox{0.73}{
\begin{tabular}{cccccccccccccc}%四个c代表该表一共四列，内容全部居中
\toprule%第一道横线
\multirow{2}{*}[-0.5ex]{\bf Method} 
& \multirow{2}{*}[-0.5ex]{\bf Lr}
%&\multirow{2}{*}[-0.5ex]{\shortstack{Time per \\ Epoch($s$)}}
%&\multirow{2}{*}[-1ex] {MU(\%)}
%&\multirow{2}{*}[-1ex] {\shortstack{Throughput \\ (Samples/Sec)}}
&\multicolumn{2}{c}{\bf ImageNet}
&\multicolumn{2}{c}{\bf CIFAR-100}
&\multicolumn{2}{c}{\bf CIFAR-10}
&\multicolumn{2}{c}{\bf iNaturalist 2018}
&\multicolumn{2}{c}{\bf iNaturalist 2019}
%&\multirow{2}{c}[-0.5ex]{Inference time}\\
&\multicolumn{1}{c}{\bf Time}\\
\cmidrule(lr){3-4} \cmidrule(lr){5-6} \cmidrule(lr){7-8} \cmidrule(lr){9-10} \cmidrule(lr){11-12}
% &&&&&\multicolumn{2}{c}{$ACC(43\%)$}&\multicolumn{2}{c}{$ACC(45\%)$}&\multicolumn{2}{c}{$ACC(43\%)$}&\multicolumn{2}{c}{$ACC(45\%)$}\\
% \cmidrule(lr){6-7} \cmidrule(lr){7-8} \cmidrule(lr){9-10} \cmidrule(lr){11-12}
&&ACC5(\%)&ACC1(\%)&ACC5(\%)&ACC1(\%)&ACC5(\%)&ACC1(\%)&ACC5(\%)&ACC1(\%)&ACC5(\%)&ACC1(\%)&Infer($s$)
\\
\midrule%第二道横线
\multirow{2}{*}[-1ex]{Primal}&$r1$&81.29&57.23&83.53&54.29&99.17&81.34&53.05&29.61&75.71&45.41&\multirow{2}{*}[-1ex]{47}\\ 
\cmidrule(lr){2-12}
%&l2&\multirow{2}{*}[1ex]{84.28}&\multirow{2}{*}[-0.5ex]{62.42}&76.29&45.60&98.35&74.89&46.00&24.80&72.07&42.27&-\\
&$r2$&{84.61}&{61.84}&81.69&50.72&99.14&80.37&45.77&23.29&73.39&43.20\\
\cmidrule(lr){1-13}
\multirow{2}{*}[-1ex]{Hyper}&$r1$&87.93&67.93&86.19&58.73&99.14&81.42&63.19\textbf{↑}&38.84\textbf{↑}&80.00&50.56&\multirow{2}{*}[-1ex]{47}\\ 
\cmidrule(lr){2-12}
%&l2&\multirow{2}{*}[1ex]{84.28}&\multirow{2}{*}[-0.5ex]{62.42}&76.29&45.60&98.35&74.89&46.00&24.80&72.07&42.27&-\\
&$r2$&{90.43}&{71.17}&84.64&55.66&99.30&82.02&58.33&34.52&79.27&49.03\\
\cmidrule(lr){1-13}
\multirow{2}{*}[-1ex]{Flatten}&$r1$&93.13&75.64&79.71&49.52&98.45&76.85&55.29&32.96&76.99&47.98&\multirow{2}{*}[-1ex]{48}\\ 
\cmidrule(lr){2-12}
%&l2&\multirow{2}{*}[1ex]{84.28}&\multirow{2}{*}[-0.5ex]{62.42}&76.29&45.60&98.35&74.89&46.00&24.80&72.07&42.27&-\\
&$r2$&{84.28}&{62.42}&76.29&45.60&98.35&74.89&46.00&24.80&72.07&42.27\\
\cmidrule(lr){1-13}
\multirow{2}{*}[-1ex]{Hydra}&$r1$&89.15&69.20&79.00&47.60&98.73&76.25&58.27&33.36&83.92&52.10&\multirow{2}{*}[-1ex]{48}\\ 
\cmidrule(lr){2-12}
&$r2$&91.39&73.36&68.14&36.51&97.60&70.60&14.02&4.95&80.42&51.25\\
\cmidrule(lr){1-13}
\multirow{2}{*}[-1ex]{ViT-Standard}&{$r1$}&{95.38}&{80.04}&91.06&66.70&99.54&86.89&\textbf{63.03}&\textbf{38.83}&\textbf{81.28}&51.12&\multirow{2}{*}[-1ex]{\textbf{50}}\\ 
\cmidrule(lr){2-12}
&{$r2$}&{\textbf{95.38}}&{\textbf{80.04}}&\textbf{91.90}&\textbf{67.56}&\textbf{99.64}&\textbf{89.07}&60.85&35.94&80.46&\textbf{52.04}&\\
\cmidrule(lr){1-13}
\multirow{2}{*}[-1ex]{ViT-Standard-Flash2}&$r1$&{95.38}&{80.04}&91.06&66.70&99.54&86.89&63.03&38.83&81.28&51.12&\multirow{2}{*}[-1ex]{46}\\ 
\cmidrule(lr){2-12}

&$r2$&{95.38}&{80.04}&91.90&67.56&99.64&89.07&60.85&35.94&80.46&52.04&\\
\cmidrule(lr){1-13}
\multirow{2}{*}[-1ex]{Vit-Standard-Ours-Simi}&{$r1$}&{93.50}&{77.38}&{90.35}&65.62&99.52&86.78&62.26&38.07&80.50&51.35&\multirow{2}{*}[-1ex]{47}\\ 
\cmidrule(lr){2-12}
&{$r2$}&{94.52}&{79.29}&91.48&66.83&99.59&88.73&59.83&35.11&80.03&50.50\\
\cmidrule(lr){1-13}
%\multirow{2}{*}[-1ex]{Vit-base-flash-opt}&l1&{95.38}&{80.04}&91.06&66.70&99.54&86.89&63.03&38.83&81.28&51.12&\multirow{2}{*}[-1ex]{44}\\ 
%\cmidrule(lr){2-12}
%&l2&{95.38}&{80.04}&91.90&67.56&99.64&89.07&60.85&35.94&80.46&52.04\\
%\cmidrule(lr){1-13}
\multirow{2}{*}[-1ex]{Vit-Standard-Ours-Flash}&$r1$&{93.50}&{77.38}&90.35&65.62&99.52&86.78&62.26&38.07&80.50{↑}&51.35\textbf{↑}&\multirow{2}{*}[-1ex]{42\textbf{↑}}\\ 
\cmidrule(lr){2-12}
&$r2$&{94.52\textbf{↑}}&{79.29\textbf{↑}}&91.48\textbf{↑}&66.83\textbf{↑}&99.59\textbf{↑}&88.73\textbf{↑}&59.83&35.11&80.03&50.50\\
\bottomrule%第四道横线
 \end{tabular}}
\end{table*}

\if 0
In this part, we demonstrate how different block partition and sampling ratios affect the attention matrices. Figure~\ref{fig:simility_chunk} provides a visual comparison through heatmap representations. 

To explore the effects of block-wise similarity on attention mechanisms, we compare the traditional softmax of the ${\mathbf{Q}\mathbf{K}^T}$  product with a novel approach fusing and sampling columns. In the case of Figure~\ref{fig:simility_chunk} (c), we first divide the $Q$ into token vectors of length 2. For each vector $\mathbf{Q_i} i\in [0, 2]$, we compute the similarity between its columns and then sample one column from each pair of similar columns. The $\mathbf{P_i}$ is obtained by multiplying the matrix $\mathbf{Q_i}$ with the fused colunm-wise matrix $\mathbf{P}$. (b) and (d) show the computed matrices $\mathbf{P}$ for chunk sizes of 1 and 3, respectively.
\fi
%-------------------------------------------------------------------------------

\if 0
From the $\mathbf{P}$ matrices computed with different chunk sizes, it can be observed that using a column similarity method based on the distributive property of matrix multiplication significantly reduces errors. Furthermore, when block-wise similarity is applied, the model's error does not accumulate as the number of input tokens increases.
\fi

\begin{figure*}[t]
\centering
\includegraphics[width=1\linewidth]{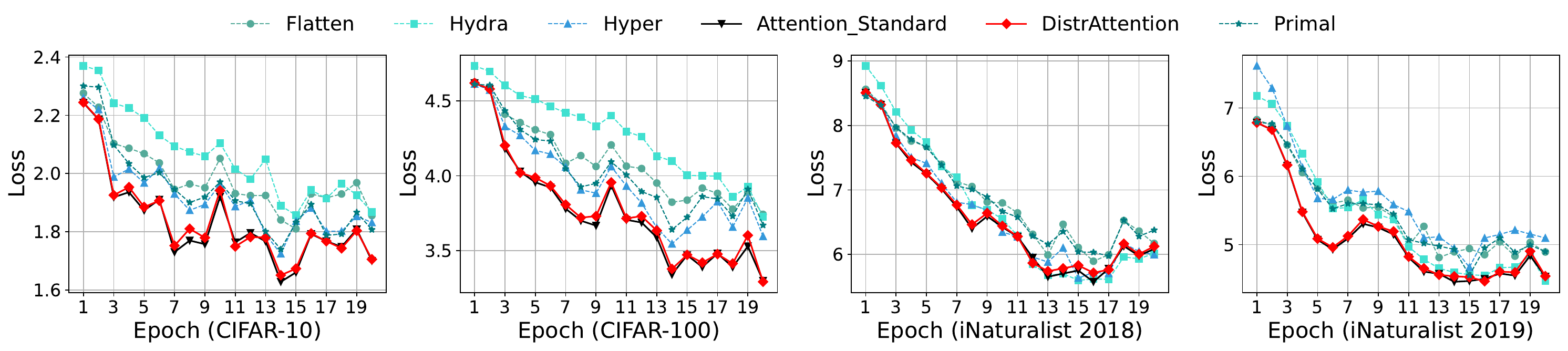}
\caption{Loss comparison during ViT model fine-tuning with different attention optimization methods, illustrating the impact on model performance and convergence.}
\label{fig:Loss_vit}
\end{figure*}

\subsection{Fine-tuning on Visual Models}
In this experiment, we evaluate attention mechanisms on the ViT model using full-parameter fine-tuning\footnote{{\tt vit\_base\_patch224} is used for fine-tuning}. The target models with different attention mechanisms are fine-tuned with the training set in datasets. All the models are trained for 20 epochs. We use a cosine learning rate scheduler and set the initial epoch as 0 and 180 for different learning rates $r1$ and $r2$. We then evaluate the accuracy and the inference time. For the accuracy, we measure {\it ACC1} and {\it ACC5}, which represent how many tasks in the test set, the ground truth appears in the top-1 and top-5 inference results. For the inference time, we measure the wall time used to complete all tasks in the test set. 

The results are reported in Table~\ref{tab:vis_method}, where {\it Primal}, {\it Hyper}, {\it Flatten}, and {\it Hydra} are approximate attention mechanisms. {\it ViT-Standard} means the default attention used in ViT. We replace the basic attention with Flash2 in {\it ViT-Standard-Flash2}. In {\it ViT-Standard-Ours}, the basic attention is changed to DistrAttention, and we set $l$ to 64 and $G^*$ to 2. {\it ViT-Standard-Ours-Flash} incorporates both our method and Flash2. It is worth noting that we do not need fine tune {\it ViT-Standard} and {\it ViT-Standard-Flash2} on ImageNet because they are exact attention. But on the other four datasets, to adapt the head of classifier at the last layer, we also fine-tune the models for 20 epochs. We report the inference time of ImageNet only due to the limited space, and other datasets present a similar performance. 

We could observe that on ImageNet, DistrAttention achieves ACC1 of 79.29\% and ACC5 of 94.52\%, which is the highest among approximate attention mechanisms. Even compared to the exact attention, the accuracy is lowered by $<$1\%. As for the inference time, our method reduces 4 seconds from the standard attention, which is the fastest among approximate attention methods. When integrated with FlashAttention, our method is the fastest one of all methods that accelerates the inference time of standard attention by 16\%. On the other datasets, similar performance trends are observed. The CIFAR-10 dataset achieves the best accuracy no matter what model is evaluated. This is because only 10 categories are trained in the dataset, and 20 epochs are enough for all methods. In contrast, the iNaturalist 2019 dataset contains 8,142 categories, which introduces more categories to the last layer. The limited fine-tuning epochs could hardly train the new parameters effectively, resulting in low accuracy. 

% across different fine-tuning methods. However, the iNat-comp2018 dataset exhibites the worst accuracy rates. This discrepancy arises because the original pre-trained model was trained for image classification with 1000 categories, whereas the CIFAR-10 dataset involves training with only 10 categories. During fine-tuning, the model can leverage and adapt the existing classification weights in the final layer, allowing for better accuracy on CIFAR-10 through various fine-tuning strategies. In contrast, the iNat-comp2019 dataset contains 8142 categories. During training, the final layer introduces new parameters, and the limited number of epochs and images is insufficient to train these new parameters effectively, resulting in low accuracy.

% Our method is the fastest and the most accurate approximate attention mechanisms while only reducing the accuracy by a small margin. By combining FlashAttention-2 and our method, the inference time is further improved.

The experimental results also reflect that DistrAttention could be integrated into pre-trained models with little efforts. Unlike other approximate attentions, DistrAttention reduces elements across the $d$ dimension of $\mathbf{Q}$ and $\mathbf{K}$, which means that neither the output shape nor token number or position is changed. We also do not introduce additional parameters into the model. In Primal, which substantially alters the attention of the pre-trained model, the poorest accuracy is achieved in only 20 epochs of fine-tuning.

Hyper achieves higher accuracy than Primal and Flatten. It rearranges the $\mathbf{Q}$ and $\mathbf{K}$ matrices by sorting them and then dividing these large matrices into smaller sub-matrices for multiplication according to the desired block size. Although this method loses information from the original tokens, it allows for complete reuse of previously learned weights during fine-tuning, thereby maintaining relatively good accuracy. In contrast, Hydra, while not introducing new parameters, alters the original attention mechanism, leading to varying accuracy rates when fine-tuning on different datasets. % Primal-Attn, however, introduces new parameters and fundamentally changes the attention mechanism, resulting in poorer performance across all datasets.

Due to the similarity of columns in Q-blocks based on the distributive property of matrix multiplication, we are able to perfectly integrate the weights from the pre-trained model without introducing new weight parameters. Consequently, after fine-tuning, our DistrAttention method maintains an error margin within 1\% compared to the original attention mechanism. In terms of inference, Hydra and Primal eliminate the need for the attention matrix $\mathbf{S}$, thereby improving inference speed compared to the original attention mechanism. However, these methods also introduce I/O bottlenecks and distribute their processes across multiple CUDA kernels, resulting in multiple data transfers. Hyper and Flatten accelerate inference by reducing data volume through sampling but cannot consolidate the entire process into a single CUDA kernel.

\begin{table}[t]
\centering
\caption{Inference Time(s) of LLama3-1B at Different Token Lengths.}
\label{tab:llm_time}
\begin{tabular}{ccccc}
\toprule
{\bf Method} & $n=256$ & $n=512$ & $n=1024$ & $n=2048$ \\ 
\midrule
{Flatten} & {0.30} & {0.31} & {0.32} & {0.35}\\ 
{Primal} & {0.27} & {0.33} & {0.41} & {0.60}\\ 
{Hydra} & {0.18} & {0.18} & {0.19} & {0.20} \\ 
{Hyper} & {0.18} & {0.18} & {0.19} & {0.20} \\ 
{Flash2} & {0.17} & {0.17} & {0.18} & {0.20} \\ 
{Attn-Standard} & {0.18} & {0.19} & {0.20} & {0.20} \\ 
\midrule
{Ours} & {0.17} & {0.17} & {0.18} & {0.19} \\ 
%{Ours-Flash} & {0.17} & {0.17} & {0.18} & {0.18} \\ 
{Ours-Flash-Opt} & {0.17} & {0.17} & {0.17} & {0.17} \\ 
\bottomrule
\end{tabular}
\end{table}

Figure~\ref{fig:Loss_vit} illustrates the loss curves of different baseline methods over 20 epochs of training across various datasets. The x-axis represents the epoch number, while the y-axis indicates the loss. 
%Each plot includes six curves representing a comparative analysis of distinct approaches. Among all dataset loss plots,
From the figure, it is observed that our loss curve variation closely aligns with the standard attention method, showing that DistrAttention is a promising approximation to the standard attention. Moreover, among all approximate attention mechanisms,  DistrAttention also achieves the lowest loss value. This observation further confirms that our method reaches low accuracy loss and effective convergence in model training.

\if 0
\begin{table}[t]
\centering
\caption{The Accuracy of Fine-Tuned LLama3-1B at Different Token Lengths.}
\label{tab:llm_acc}
% \renewcommand{\arraystretch}{1.5}
\begin{tabular}{ccc}
\toprule
\bf{Method} & \bf{$n=256$} & \bf{$n=512$} \\ 
\midrule
Flatten-Attn & 31.25 & 31.27\\ 
Primal-Attn & 24.11 & 31.25\\ 
Hydra-Attn & 29.46 & 22.75 \\ 
Hyper-Attn & 33.29 & 32.50 \\ 
Flash-Attn & 33.29 & 32.50 \\ 
Base-Attn & 33.29 & 32.50\\ 
\midrule
Ours & 31.37 & 31.50 \\ 
\bottomrule
\end{tabular}
\end{table}
\fi

\begin{table}[t]
\centering
\caption{The Accuracy of Fine-tuning LLama3-1B at Different Token Lengths.}
\label{tab:llm_acc}
\resizebox{1\linewidth}{!}{
\begin{tabular}{cccccccc}
\toprule
\bf{Method} & Flatten & Primal &Hydra & Hyper &Flash2 &Attn-Standard &Ours \\ 
\midrule
\bf{$n=256$} &31.25& 24.11& 29.46& 33.29& 33.29& 33.29& 31.37\\
\midrule
\bf{$n=512$}& 31.27& 31.25& 22.75& 32.50& 32.50& 32.50& 31.50\\
%Flatten & 31.25 & 31.27\\ 
%Primal & 24.11 & 31.25\\ 
%Hydra & 29.46 & 22.75 \\ 
%Hyper & 33.29 & 32.50 \\ 
%Flash2 & 33.29 & 32.50 \\ 
%Attn-Standard & 33.29 & 32.50\\ 
%\midrule
%Ours & 31.37 & 31.50 \\ 
\bottomrule
\end{tabular}
}
\end{table}

\begin{figure*}[t]
  \centering
  \includegraphics[width=\textwidth]{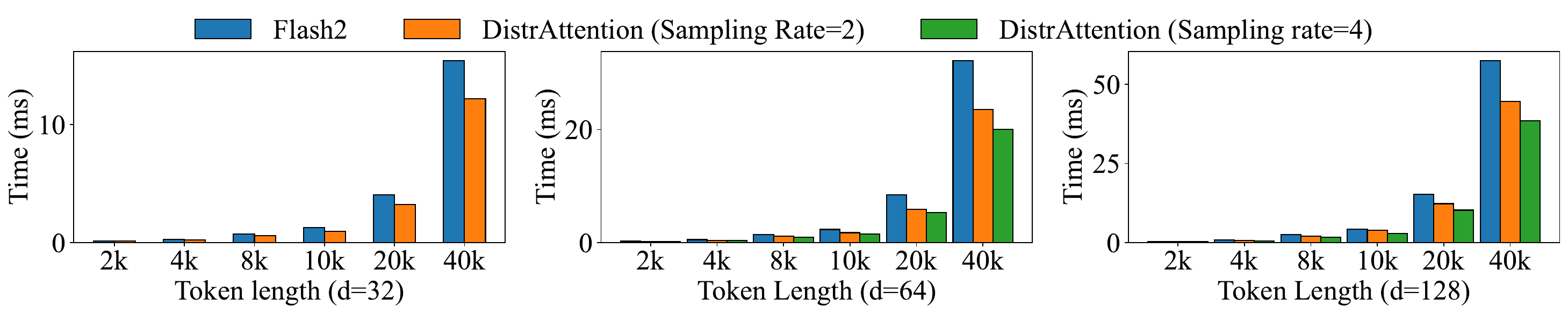}
  \caption{Comparison of time spent on computing attention between Flash2 and our approach under different $d$ and token lengths. The term $Sampling~ rate=2$ and $Sampling~rate=3$ refer to the different sampling and fusion rates used for the $Q$ and $K$ when calculating the attention matrix.}
  \label{fig:each_layer_time}
\end{figure*}
\subsection{Evaluation on Language Models}

We also evaluate various attention mechanisms on the Llama-3.1-1b pre-trained model. We fine-tune the pre-trained Llama-3.1-1B model with the MathInstruct dataset for five epochs, and use the MMLU dataset as the test set. We change the pre-filled token length and measure the Time to First Token (TTFT). The results are reported in Table~\ref{tab:llm_time}. We could observe that whatever the length of pre-filled token is, DistrAttention and DistrAttention integrated with Flash2 are always the fastest. It is interesting that as approximate attention mechanisms, Flatten and Primal take even longer inference time compared to the standard attention. This is because that these two methods introduce additional parameters into the network architecture and while the token length is small, additional time is required in the pre-filling stage. In contrast, Hyper, Hydra, DistrAttention, and Flash2 do not introduce additional parameters into the network and thus achieve faster inference time compared to the standard attention.

We set the token length to 256 and 512, and the results are shown in Table~\ref{tab:llm_acc}. The results show that our method achieves the second highest accuracy among approximate attention mechanisms, which lowers the accuracy by less than 1\% from the exact attention.  Both of the experiments demonstrate that our approach could effectively improve the inference performance and degrade the accuracy by a small margin.

\begin{table*}[t]
\caption{Accuracy and Inference Time of ViT and BERT Pre-Trained Models Without Fine-Tuning.}
\label{tab:no_train}
\centering
\begin{tabular}{c|cccc|cccc}
\toprule
& \multicolumn{4}{c|}{\bf Acc} & \multicolumn{4}{c}{\bf Times (s)}  \\
\midrule
{\bf Model}  & Attn-Standard& Hydra& Flash2  & Ours   & Attn-Standard     & Hydra    & Flash2  & Ours \\ 
ViT\_base\_patch224  & 81.1\% & 0.1\% & 81.1\% & 75.1\% & 49   & 48  & 46     & 41  \\
ViT\_base\_patch384 & 84\%   & 0.1\% & 84\%   & 77.12\% & 180  & 80  & 146    & 125 \\
bert\_neo\_uncased  &62.75\% &22.5\%&62.75\%&62.67\%&67&65&64&60 \\
\bottomrule
\end{tabular}
\end{table*}

% Our proposed method, which employs segmented parallel similarity computation to minimize errors, achieves inference performance closely approximating that of Base-Attn. This method divides the attention calculation into segments, allowing for parallel processing and error reduction, ultimately resulting in inference outcomes that are nearly equivalent to those achieved by the standard attention mechanism. By adopting this strategy, our approach not only enhances computational efficiency but also maintains high accuracy across varying token lengths, making it a viable alternative to traditional attention mechanisms in large-scale language modeling tasks.

% Tables~\ref{tab:llm_time} and~\ref{tab:llm_acc} demonstrate that our approach can maintain an error margin of approximately 1\% while improving inference time by 15\% in large-scale models. Notably, this enhancement is achieved through fine-tuning rather than training from scratch, which convincingly attests to the efficacy of our method.

\subsection{The Attention Time}
Figure~\ref{fig:each_layer_time} illustrates a comparison of the time spent on computing attention between Flash2 and our proposed method under conditions where batch size is set to 1 and the number of heads is 10, across varying token lengths and $d$. From Figure~\ref{fig:simility_chunk}, it can be observed that employing different sampling rates has a negligible impact on the attention matrix. Therefore, in this section, we utilize two sampling rates to examine their computational time across varying token lengths. As clearly demonstrated in Figure~\ref{fig:each_layer_time}, our proposed method outperforms Flash2 irrespective of the chosen sampling rate or token length. Notably, when $d=32$, there is no instance of a sampling rate equal to 4 compared to the cases where $d$ equals 64 or 128. This omission is due to the fact that at $d=32$, applying a sampling rate of 4 for the sampling and fusion process of $\mathbf{Q}$ and $\mathbf{K}$ matrices results in a column dimension of 8. Such a scenario does not lend itself to acceleration via tensor core computations, which are optimized for higher dimensions. Consequently, the sampling rate of 4 is excluded for $d=32$ to ensure compatibility with tensor core optimizations and maintain computational efficiency.

In the section, across the three different dimensions $d$, a common observation can be made: when the token length is small, our method exhibits comparable computational times to Flash2. However, as the token length increases, the time discrepancy between our method and Flash2 becomes increasingly pronounced. This phenomenon can be attributed to the underutilization of GPU internal resources at shorter token lengths, leading to similar computation times for both methods. Conversely, with an increase in token length, threadblocks, which are directly proportional to token length, fully occupy the tensor cores on each SM.

The superior performance of our approach, especially as token length increases, can be explained by the effective utilization of different sampling rates. By employing various sampling rates, our method reduces the tensor core computation time required for calculating the attention matrices post-tile segmentation. This optimization allows for more efficient use of the tensor core resources, which would otherwise be a limiting factor due to their high demand. Specifically, different sampling rates enable a reduction in the computational load for each individual operation, thereby allowing the tensor cores to process a greater number of matrices within the same timeframe. This strategy not only maximizes the throughput of tensor cores but also ensures that the limited hardware resources are used more effectively, resulting in significant performance gains over Flash2 as token lengths become longer.

Thus, our method demonstrates a marked improvement in efficiency and speed, particularly advantageous in scenarios involving extensive token lengths. By enabling the computation of a greater number of attention matrices on the same tensor cores, our method achieves a 37\% improvement over Flash2.

\subsection{Analysis without Fine-Tuning}
%Table~\ref{tab:no_train} illustrates the performance of various attention optimization methods applied to pre-trained ViT and BERT models in terms of inference time and accuracy. For the Vit\_base224 and Vit\_base384 models, evaluation is conducted using the test set from the ImageNet dataset, with the difference between these models being the input image sizes. The bert\_neo\_uncased pre-trained model's performance is assessed using the Political Language Part-of-Speech Classification dataset for each method.

We also evaluate our method and baselines on pre-trained models without fine-tuning. Among approximate baselines, we select Hydra because it does not introduce new parameters into the model. The experiments are carried out on {\tt vit\_base\_patch224}, {\tt vit\_base\_patch384}, and {\tt bert\_neo\_uncased}. We measure both the accuracy and the inference time, and the results are shown in Table~\ref{tab:no_train}.

For the exact attention like the standard attention and Flash2, the accuracy is the highest. Flash2 reduces the inference time of different models by 6\%, 19\%, and 4\% without compromising the accuracy. On the other hand, our method further reduces the inference time by 16\%, 31\%, and 12\%. The price is that the accuracy is reduced by 6.00\%, 6.88\%, and 0.08\%. This indicates that for specific models, even without the fine-tuning step, our method could improves the inference performance with little degradation of the accuracy.

From the results, Hydra exhibits significantly lower accuracy on the two ViT models. This is attributed to that in large models, the inference result relies on the attention scores of pairwise tokens while Hydra eliminates the attention matrix in its design. Without fine-tuning, it fails to adapt the parameters to the new model. In the BERT model, the relatively higher accuracy is due to the limited number of part-of-speech categories and the tendency for most POS tags falling into a single predominant category.

\begin{table}[t]
  \centering
  \caption{The time ($ms$) of executing Flash2 and Ours with varying $GPU$.}
  \label{tab:Mutil-gpu}
  \begin{tabular}{ccccc}
    \toprule
    \bf{Method} & $GPU$=1 & 2 & 4 &\\ 
    \midrule
    Flash2 & 1299 & 1768 & 1471 & \\ 
    Ours & 846 & 1361 & 1359 & \\
    \bottomrule
  \end{tabular}
\end{table}

\subsection{The Analysis of Multi-GPU Experiments}
Distributed training and inference have become mainstream tasks in current research and applications. To evaluate the effectiveness of DistrAttention in multi-GPU scenarios, we compared the time consumption of DistrAttention with that of Flash2 across different numbers of GPUs. We randomly construct $\mathbf{Q}$, $\mathbf{K}$, and $\mathbf{V}$ matrices ($H = 480$, $N = 20480$, $d = 128$), where an element is individually generated following a uniform possibility in (0, 1). Then, we calculate the time consumption of DistrAttention and Flash2. The experiment is repeated 30 times, and the results are recorded as the average time. In the experiment, we set the block size to 128 and the sampling rate to 2. To eliminate the impact of inter-GPU communication on the total time, we choose not to evenly distribute the $\mathbf{Q}$, $\mathbf{K}$, and $\mathbf{V}$ matrices across different GPUs. Instead, we split them into multiple blocks with $H = 20$ and scatter these blocks to different GPUs in multiple rounds. Then, we implemented a double-buffering system to facilitate the transmission of the next data chunk, and by designing the parallelization of computation and data flow, we achieve overlap between computation and data transmission. With the configurations, we calculate the consumption time. 

Table~\ref{tab:Mutil-gpu} shows the time spend by Flash2 and DistrAttention in distributed computation for calculating the same attention, which includes the number of the GPUs ({\it GPUs}) used for attention computation and the comparison methods. The results demonstrate that DistrAttention achieves notable acceleration in both single-GPU and multi-GPU settings, and the acceleration is more pronounced in the single-GPU scenario, reaching up to 34.87\%, due to the absence of data scatter and gather processes. In the multi-GPU distributed scenario for computing attention, our approach achieves a speedup of at least 7.61\% to 23.02\% compared to Flash2. DistrAttention achieves significant acceleration in both single-GPU and multi-GPU scenarios. This experiment further demonstrates the flexibility and robustness of our method. 

\subsection{The Analysis of LSH-Based Grouping}
Grouping $\mathbf{Q}$ based on LSH is an indispensable part of DistrAttention, and the performance of this component directly impacts the overall efficiency of DistrAttention, therefore, we design an experiment to evaluate the effectiveness of this component. We measure the computation time of this component for $\mathbf{Q}$ with varying dimensions. We randomly construct $\mathbf{Q}$ matrices ($N, d = 128$), where an element is individually generated following a uniform possibility in $(0, 1)$. In the experiment, $N$ is selected among 2048, 4096, 20480, and 40960. Then, we repeat this experiment 100 times, and calculate the average time of Grouping $\mathbf{Q}$ with different $N$. The LSH-based grouping time for $N$ sizes 2048, 4096, 20480, and 40960 is 0.14$ms$ (74.8\%), 0.14$ms$ (37.8\%), 0.15$ms$ (4.1\%), and 0.15$ms$ (1.3\%). When $N$ = 2048 and 4096, it can be observed that the LSH-based grouping accounts for a relatively higher proportion of overall computation time compared to matrix calculations. Because the computation time for small matrices is dominated by the overhead of launching CUDA kernels, which is approximately 0.1$ms$. The results demonstrate that this lightweight LSH-based grouping method has minimal impact on the performance of DistrAttention.
\section{Related Work}
\label{sec:related_work}
%-------------------------------------------------------------------------------
% The attention~\cite{vaswani2017attention} architecture plays a crucial role in both vision and language models through its powerful self-attention mechanism. In vision tasks, Attention can capture global dependencies between different regions of an image and process various parts of the image in parallel, thereby enhancing the model's ability to understand complex image structures. In NLP, document-level processing necessitates working with extensive articles~\cite{pappagari2019hierarchical,kwiatkowski2019natural}, and the effectiveness of language models tends to improve with longer input sequences. In computer vision, various tasks, such as those involving high-resolution images, convert these images into long sequences of patches before the transformer model analyzes them. However, the increasing demands for processing longer input sequences have resulted in a quadratic rise in execution time, primarily driven by the exponential growth in I/O operations. As the length of the input sequences grows, the number of I/O operations required to handle and process the data increases significantly, leading to a substantial increase in computational overhead. Therefore,  there has been a proliferation of research dedicated to optimizing computer vision and NLP transformer architectures.

Optimizations of the attention layer have been studied extensively in the past, to address the performance issue of the attention mechanism under long context length. Among these proposed approaches, Flash Attention~\cite{dao-2022-flash, dao-2024-flash} is one of the most widely used optimizations, where attention computation is divided into blocks and fused to reduce the need to swap data in and out of global memory, and therefore improving overall performance. It also demonstrates the importance of carefully managing memory layout and access, where works like Paged Attention~\cite{kwon2023efficient} has proved its effectiveness during LLM inference, especially for KV Caches, which is also component that attracts researchers' attention for optimization~\cite{paszke2019pytorch,olston2017tensorflow,yu2022orca,fasttransformer,zhang2023h2o}. 

Besides, researchers have also turned to approximate attention and/or linear attention to further improve performance of the attention layer. For example, Flatten Transformer~\cite{han2023flatten} proposes Focused Linear Attention, where a mapping function and an efficient rank restoration module is proposed to address the ability to focus and feature diversity without using softmax function in the attention layer and without introducing high computation complexity. Primal Attention~\cite{chen2024primal} is another work to improving efficiency of the self-attention mechanism via asymmetric Kernel Singular Value Decomposition, where the authors formulate a primal-dual representation of self-attention to facilitate a low-rank approximation of the attention matrix. Hyper Attention, proposed in ~\cite{han2023hyperattention}, claims to achieve sub-quadratic time complexity for the attention layer with regards to the context length, where locality sensitive hashing (LSH) is used to identify large entries and fast approximate matrix multiplication is performed by sampling. The LSH-based approaches are also proposed in Reformer~\cite{kitaev2020reformer} and KDEFormer~\cite{zandieh2023kdeformer} to exploit the sparsity of the attention matrix. Similarly, by employing Random Feature Mapping and Low-Rank Approximation, Performer~\cite{choromanski2024learning} and Linformer~\cite{wang2020linformer} address the quadratic computational bottleneck associated with standard self-attention mechanisms, enabling more efficient processing of long sequences. Despite the promising theoretical result of various linear attention mechanism, it is challenging to translate it into piratical performance advantages due to I/O bottleneck and cumsum limit in causal settings. Lightning Attention\cite{qin2024transnormerllmfasterbetterlarge, qin2024lightning} is an attempt to address this, where the authors claim significant performance advantage of the linear attention, even in causal settings, by handling the computation of intra-block and inter-block components separately in linear attention calculation, coupled with careful tiling design to leverage the full capability of specific GPUs. Hydra Attention~\cite{bolya2022hydra}, on the other hand, is focused on the multi-head attention mechanism. In addition, ~\cite{team2024gemma} constructs each group's key and value heads by performing mean pooling on all the original heads within that group. Primer-attn~\cite{so2021searching} employs convolutional operations to merge information from multiple attention heads, effectively reducing the attention mechanism's I/O overhead.

\section{Conclusion}
Despite the unique self-attention mechanism, efficient parallel computation capabilities, and superior sequence modeling performance that have made the transformer architecture indispensable in natural language processing (NLP) and beyond, its application to long-sequence tasks is constrained by a time complexity that scales quadratically with token length $O(n^2)$. To address this limitation, we achieve significant improvements by partitioning the $\mathbf{Q}$ and $\mathbf{K}$ matrices, implementing parallel sorting, sampling, and merging similar columns to reduce the embedding dimensionality $d$. These strategies are integrated with Flash Attention techniques, optimizing the chunk size of $\mathbf{Q}$, $\mathbf{K}$, and $\mathbf{V}$ matrices through a refined balance among tensor cores, shared memory, and threadblock. This approach leads to higher accuracy and faster inference speeds compared to existing state-of-the-art attention optimization methods across various models.

%% Reference
\balance
%\newpage
\bibliographystyle{ACM-Reference-Format}
\bibliography{attention}


%\begin{acks}

%\end{acks}

\end{document}